# Self-Selective Correlation Ship Tracking Method for Smart Ocean System

Xu Kang, Bin Song, Jie Guo, Xiaojiang Du and Mohsen Guizani

**Abstract:** In recent years, with the development of the marine industry, navigation environment becomes more complicated. Some artificial intelligence technologies, such as computer vision, can recognize, track and count the sailing ships to ensure the maritime security and facilitates the management for Smart Ocean System. Aiming at the scaling problem and boundary effect problem of traditional correlation filtering methods, we propose a self-selective correlation filtering method based on box regression (BRCF). The proposed method mainly include: 1) A self-selective model with negative samples mining method which effectively reduces the boundary effect in strengthening the classification ability of classifier at the same time; 2) A bounding box regression method combined with a key points matching method for the scale prediction, leading to a fast and efficient calculation. The experimental results show that the proposed method can effectively deal with the problem of ship size changes and background interference. The success rates and precisions were higher than Discriminative Scale Space Tracking (DSST) by over 8 percentage points on the marine traffic dataset of our laboratory. In terms of processing speed, the proposed method is higher than DSST by nearly 22 Frames Per Second (FPS).

**Keywords:** Ship tracking; correlation filter; negative samples mining; self-selective model; box regression

## 1. Introduction

The ocean is rich in resources, which will provide tremendous materials for the shortage of human resources. However, due to the imperfect infrastructure, the development and utilization of marine resources are still in its infancy. With the development of the marine industry, marine transportation needs large-scale management, marine resources sharing and marine activities need comprehensive coordination, which can create new wealth for the marine field. Smart Ocean is a comprehensive perception and understanding of ocean data, and then provides intelligent interactive services. Smart Ocean is a good combination of information technology and traditional ocean industry. The work of Smart Ocean includes the links of marine equipment, human activities, marine environment, and management subjects. The construction of the Smart Ocean can be classified into four categories: perception network, communication network, artificial intelligence platform, and application groups. Perception networks mainly include large-scale sensors installed at sea, ships, ports or bays, which transmit captured data to cloud platforms through communication networks. The collected data can be analyzed and processed by the artificial intelligence algorithm on the cloud platform, and the results can be fed back to application groups. The captured image information can be processed by artificial intelligence technology, which can identify, track and count the ships in real time. By tracking the course of the ships sailing into or out of the bay, the relative position of the ships and the restricted area can be computed to analyze the ships' behavior.

Because of its cross-disciplinary, heterogeneous data formats and vulnerability to weather and environmental disturbances, it is very difficult to track ships at sea level. The core issue of target tracking in smart ocean system is how to select an effective feature extraction method for different scenes to represent the target ship. Single ship tracking task is to predict the size and location of the target in the next frames given the size and location of the target ship in the initial frame of a video sequence. Feature engineering transforms the data of the raw pixel domain into abstract features, which can describe the appearance of the ship and make the tracker perform better on the learned model than on the original pixels. In addition to feature engineering, the quality of pattern recognition (PR) system depends significantly on the machine learning (ML) algorithm model used by the smart ocean system itself. There are various kinds of feature extraction methods in the pattern recognition system. When combined with different types of trackers, it is difficult to grasp which feature extraction method is most effective. Multi-feature fusion and multi-task joint learning can solve this problem well. By using different kinds of features in different trackers, the model can analyze the reliability of each feature and learn adaptively. In addition, traditional target tracking methods need to extract multi-scale features when dealing with target scale transformation and even need to train a scale filter especially, which is very time-consuming. The target aspect ratio can only be adjusted according to a certain proportion, and it cannot cope with the deformation of the target well.

In this article, aiming at the scaling problem and boundary effect problem of traditional correlation filtering methods, we propose a self-selective correlation filtering method based on box regression (BRCF) for Smart Oceans. The method mainly includes 1) A hard negative samples mining method to reduce the boundary effect of the correlation filter. 2) A self-selective model with adaptive multi-response fusion method to strengthen the classification capability of the tracking model. 3) A key points matching method for scale pre-estimation of the tracking target. 4) A bounding box regression method to adjust the bounding box of the tracking target freely.

The rest of the article is organized as follows: Section 2 introduces the traditional tracking algorithms for visual tracking. Section 3 describes a self-selective model with adaptive multi-response fusion and negative samples mining method. A scale pre-estimation method using points matching and a bounding box regression method for scale fine-tuning is presented in Section 4. Section 5 compares the success rate, precisions, and speed of the proposed method with other two methods based on our traffic dataset. We conclude this article in section 6.

## 2. Related Works

The traditional target tracking methods are usually based on a generative model and a discriminant model. Through mathematical modeling, the generative model represents the distribution of image data from a statistical point of view, which can reflect the similarity of between the pixel features and the searching area and find the most reliable position.

Since 2010, target tracking methods based on Correlation Filter (CF) have been springing up and achieved good tracking results. In 2010, Bolme et al. firstly used CF for target tracking [1]. The image signal was transformed into the frequency domain through Discrete Fourier Transform (DFT), and the correlation between signals was measured by minimum mean square error (MMSE). In [2-4], the authors used the property between the circular matrix and DFT to further simplify the operation of the cyclic shift operation. Galoogahi et al. proposed a CF tracking method based on the multi-channel feature map so that the CF can be applied to the RBG image [5]. Henriques et al. used the regularized least-squares classifier (RLSC) to optimize the mean square error (MSE) between the signals [6]. The kernel method was first used to solve the RLSC with signals in high dimensional space which increased the speed of the CF method. Danelljan used CF to solve the scale transformation [7]. The two-dimensional CF and the one-dimensional CF were combined together and can be updated at the same time. Ma et al. add a classifier to detect the target confidence, which can correct the tracking errors [8]. Galoogahi et al. used the larger size of image blocks to find the object location but used the smaller size of filters to raise the proportion of positive samples [9].

Many methods use regularization term to evaluate the spatial information around the tracking target, so as to improve the tracking performance [10, 14, 15, 17, 18, 21, 22]. In order to reduce the boundary effect, a spatial regularization method was described to penalize correlation filter coefficients according to their spatial location [10]. Lukezic et al. utilized channel reliability to promote the robustness of features and used spatial reliability to reduce the boundary effect [17]. Mueller et al. considered more background patches as context-aware information and took more negative samples to train the model [15]. A fully convolutional neural network with spatially regularized kernels was presented to analyze the spatial information of deep features so that the filter kernels can focus on a specific region of the target [14]. He et al. also exploited the discriminative power in the CNN features by summing the weighted convolution responses from each feature block to produce the final confidence score [18]. Sun et al. adopted an alternating direction method to solve a CF-based optimization problem which jointly models the discrimination and reliability information [21]. Li et al. proposed a spatial-temporal regularized correlation filters which handles the boundary effects without much loss in efficiency and achieve superior performance [22]. Several works used multi-task or multi-feature fusion to enhance the robustness of trackers [11, 13, 20]. Bertinetto et al. utilized color statistics as complementary traits to make the model cope with the situations exhibiting motion blur and illumination changes [11]. Zhang et al. first presented the multi-task correlation filter combining with a particle filter [13]. Wang et al. proposed an efficient multi-cue analysis framework which explores the context among the adopted multiple features and the strength of them, and each expert in the model tracks the target independently [20]. A novel large margin object tracking method with a multimodal target detection technique is proposed to improve the target localization precision in [16], and feedback from high-confidence tracking results was adopted to avoid the model corruption problem and model drift. Danelljan et al. exploited a factorized convolution operator in the compact generative model of the training sample distribution, which drastically reduces memory and time complexity [12]. Wang et al. propose to generate hard positive samples via adversarial learning for visual tracking [19]. The generated diverse target object images can enrich the training dataset and enhance the robustness of visual trackers.

With the development of sensor technology and improvement of hardware, image sensors can be combined with wireless sensors in the Smart Ocean system [23, 24, 25, 26, 27, 28, 29, 30, 31, 32]. Du et al. provided security and privacy for sensor networks in terms of computation, communication, memory/storage, and energy supply [23, 24, 25, 26]. Tian et al. introduced the latest application of video surveillance in the intelligent transportation system [27]. Zhang et al. analyzed a socially aware Internet of vehicles with a deep reinforcement learning method [28]. Guo et al. introduced applications of compressive sensing (CS) in vehicular infotainment systems, including object tracking for safety [29, 30]. Chavez-Garcia and Aycard presented multi-sensor fusion for accurate object tracking for advanced driver assistance systems [31]. Zhang et al. used a multimodal data fusion method to track vehicles through multi-camera linkage [32]. Kang et al. proposed some vehicle classification, vehicle tracking and anomaly detection methods based on road traffic scenes [33, 34, 35]. A new method to extract the characteristics of ship attitude angle by using radar track information was presented in [36]. Kalman Filter (KF) based techniques for tracking ships using Global Positioning System (GPS) data were introduced in [37]. Ban et al. used a support vector regression (SVR) method for deposing outliers and regressing polluted ship tracks [38]. Chen et al. proposed a novel method based on mean shift for automatic detection and tracking of a ship in the corrected video sequences [39]. Leclerc et al. experimented with pre-trained convolutional neural networks to perform ship classification on a limited ship image dataset [40]. Xiao et al. proposed a ship tracking algorithm of harbor channel based on orthogonal particle filter employing Bayes state estimation [41].

**3. Self-selective Model with Negative Samples Mining**

The purpose of this section is to give our self-selective model based on hard negative samples mining. At the beginning of this section, we start with the principle of kernelized correlation filter (KCF) and the online updating rule of it. Three main components of the tracking method are the regularized least-squares classifier, circuit convolution of image features and the DFT. When using traditional KCF method for target tracking, the feature selection is too single to cope with the changes in the surrounding environment, and there is a serious boundary effect. In order to solve these problems, a new method of mining positive and negative samples based on local region is proposed in the following section, which reduces the complexity of a single model and improves the classification ability of the model. At the end of this section, a multi-feature adaptive selection method is proposed to improve the robustness of the tracking model under the dynamic background environment.

### 3.1. Updating rule of KCF

In machine learning, we assume that we have training dataset $S = (x_1, y_1), (x_2, y_2), \ldots, (x_l, y_l)$, including $l$ independent training samples, where $x_i \in R^l$, $y_i \in R$. $f$ is defined as a classification function. It is a convex mapping function in positive definite Hilbert space:

$$f(z) = \sum_{i=1}^{l} c_i \kappa(z, x_i) \tag{1}$$

where $c \in R^l$ indicates the parameters of the model, $x \in R^l$ is the input signal and $\kappa(\cdot)$ denotes the mapping results of circular convolution between the input signal and the sample signal in high-dimensional Hilbert space. Based on the dataset above, the loss function of the classifier is:

$$\min_{f \in H} \frac{1}{l} \sum_{i=1}^{l} \|y_i - f(x_i)\|^2 + \lambda c^T K c \tag{2}$$

where $K \in R^{l \times l}$ is defined as the kernel matrix composed of sample vectors. $K_{ij} = \kappa(x_i, x_j)$ is the element of $K$ at $(i, j)$. Gaussian function and linear function are commonly used kernel functions. Then the kernelized version of Ridge Regression is as follows:

$$c = (K + \lambda I)^{-1} y \tag{3}$$

A circular matrix is given by:

$$X = circ(x) = \begin{bmatrix} x^T \\ cshift(x^T, 1) \\ \vdots \\ cshift(x^T, l-1) \end{bmatrix} \tag{4}$$

As shown in the formula above, each row in the circular matrix $X$ is the circular shift of $x^T$ in different times. $cshift(x, i)$ circularly shifts the base signal $x$ in $i$ times. According to this definition, the classifier in formula (1) can be expressed as:

$$f(z) = circ(k^{xz})^T c \tag{5}$$

where $k^{xz}$ indicates the cross-correlation vector between $x$ and $z$. The element $k_i^{xz}$ in the kernelized self-correlation vector $k^{xz}$ is defined as $k_i^{xz} = \kappa(z, cshift(x, i))$. According to matrix theory that a circular matrix can be diagonalized by DFT:

$$X = F \operatorname{diag}(\hat{x}) F^H \tag{6}$$

where $X$ is the circular matrix of the raw signal $x$. $\hat{x}$ is the DFT form of $x$ in frequency domain, $F$ denotes the DFT matrix. $H$ indicates the complex conjugate transpose. The classifiers in the frequency domain can be expressed as:

$$\hat{f}(z) = \hat{k}^{xz} \odot \hat{c} \tag{7}$$

Substituting formula (6) in formula (3), the further derivation gives the classifier in the frequency domain:

$$\hat{f}(z) = \hat{k}^{xz} \odot \hat{c} = \frac{\hat{k}^{xz} \odot \hat{y}}{\hat{k}^{xx} + \lambda l} \tag{8}$$

where x is the training samples and z is the input signal in the testing stage. $\hat{k}^{xx}$ denotes the kernelized self-correlation vector in the frequency domain. The element $k_i^{xx}$ in the kernelized self-correlation vector $k^{xx}$ is defined as $k_i^{xx} = \kappa(x, cshift(x,i))$, where $\kappa(\cdot)$ is the kernel function. $\hat{k}^{xz}$ indicates the cross-correlation vector between x and z in the frequency domain. The element $k_i^{xz}$ in the kernelized self-correlation vector $k^{xz}$ is defined as $k_i^{xz} = \kappa(x, cshift(z,i))$. $\hat{y}$ transforms the Gaussian label vector y in the frequency domain.

In the calculation of cross-correlation kernel vectors, the circular matrix of the raw signal should be calculated by a circular shift, and then be the inner product with the input vector. By the property of Fourier diagonalization for the circular matrix, Fast Fourier Transform (FFT) is used to work out the cross-correlation kernel vectors with the transformation of the raw signal and the input signal. Then the inner product of the two signals is carried out and the Inverse Fast Fourier transform (IFFT) is employed. As a result, the computational complexity is greatly reduced when implemented by computers.

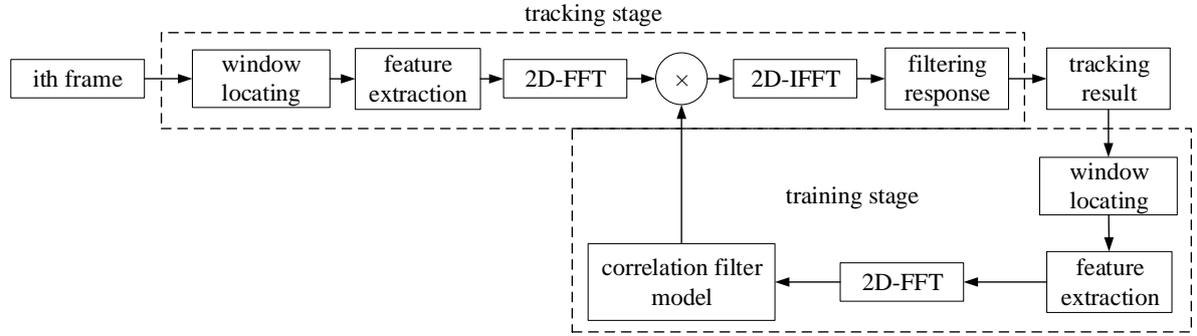

**Figure 1.** A schematic diagram of KCF. The whole process is divided into two steps: training and tracking. After updating the correlation filter model by the tracking on the last frame, the model can get the tracking result on the current frame. FFT and IFFT respectively denote the Fast Fourier Transform and the Inverse Fast Fourier Transform.

The correlation filter of one-dimensional signals can be generalized to the form of two-dimensional signals. The parameters $\hat{k}^{xx}, \hat{k}^{xz}, \hat{y}$ in formula (4) can be generalized to the self-correlation matrix, cross-correlation matrix and Gaussian matrix. $\hat{f}(z)$ in formula (4) represents the filtering response in the frequency domain, after a 2D-IFFT the filtering response in the time domain is generated. Figure 1 shows the execution process of the traditional correlation filtering algorithm. The location and size of the target object are given in the first frame. First, the HOG features of the image block are extracted. Then, the HOG feature is transformed into the frequency domain by two-dimensional DFT. In the next frames, the window is located at the position predicted in the previous frame during the tracking stage. In the training stage, the window is located at the predicted position in the current frame. The parameters of the correlation filter are calculated by combining the self-correlation kernel vector. The two-dimensional parameters of KCF can be calculated by the formula (4) directly. An iterative fashion is used to update the model:

$$\hat{c} = \begin{cases} \hat{c}_t, & t=1 \\ \hat{c} + \alpha\hat{c}_t, & t>1 \end{cases} \quad (9)$$

where $c_t$ is the two-dimensional parameters calculated in the current t'th frame, $\alpha$ is the learning rate.

The model parameters are calculated from the next frames and fine-tuned on the parameters from the first frame. The shape of the target in the first frame is very important. If the next video frame has a drastic distortion compared with the first frame, the performance of the model will be greatly reduced.

*3.2 Local region hard negative samples mining*

Figure 2 shows the general relationship between the classification hyperplane determined by the parameters and the training samples in machine learning. The white circle in the graph represents the positive samples, the dark circle represents the negative samples. The thick solid line in the middle black represents the n-dimensional classification hyperplane determined by the n-dimensional parameters of the classifier. The dashed line represents the area close to the hyperplane of classification.

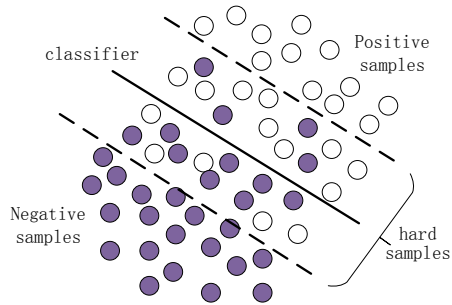

**Figure 2.** The relationship between classification hyperplane and samples. The white circles and the dark circles separately indicate the positive samples and the negative samples. The samples inside the dotted line are the hard samples which will not be classified correctly.

It is obvious from the figure that the classification hyperplane is determined by the samples near the hyperplane inside the dotted line. When classifying these samples, the classifier easily judges the positive samples (white circles) below the hyperplane as the negative samples (dark circles). To the contrary, the negative samples above the horizontal line will be classified as positive samples. So we call the samples inside the dotted line as hard samples. Therefore, if we can use the hard samples, we will save a lot of training time and improve the classification accuracy at the same time.

The HOG features at the center are moved to the origin at the upper left corner when transformed into the frequency domain. Therefore, the self-correlation kernel matrix is obtained after the feature is shifted one cycle (W and H in two directions). The two-dimensional label matrix with the same size is also obtained by the circular shift in the same range. However, when shifting to the edge of the image, the target will be split into two parts, which is called the boundary effect. In this case, the split samples are useless and time-consuming for classifiers. For the response of the circular convolution, we hope to select more samples in the middle region of the response as hard samples.

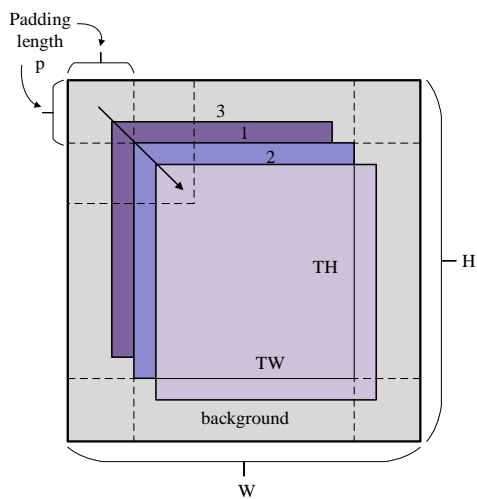

**Figure 3.** Circular samples based on the local region. The base sample is the target at the center of the image patch shown as the bounding box 1. Box 2 and 3 respectively indicate the circular shifts of the base sample. The arrow in the figure represents the direction of the circular shift.

Figure 3 shows a sampling process of the sliding samples based on the local region. The image represents the image block extracted from the target position. It is composed of the target region at the center and the background around with width P. The height of the target is TH and the width is TW. Therefore, the height of the image are H = TH + 2P, W = TW + 2P. The base sample is the target at the center of the image patch shown as the bounding box 1. Box 2 and 3 respectively indicate the circular shifts of the base sample. The background region is also shifted circularly with the target area at the same time. The shifting distance is restricted to the range $x_{shift}, y_{shift} \in [-P, P]$, corresponding to $2P \times 2P$ the region inside the dotted line. $x_{shift}, y_{shift}$ are the offsets alongside the horizontal axis and vertical axis. It is obvious from Figure 3 that the top left vertex of the target frame just slips over the range $x_{tl}, y_{tl} \in [0, 2P]$, and the bottom right vertex of the target frame just slips over the range $x_{br} \in [TW, W], y_{br} \in [TH, H]$. Circular convolution in this range can avoid the segmentation of target samples. In addition, there is a strong correlation between the base sample and the shifting samples close to it, which belongs to the hard samples to be mined.

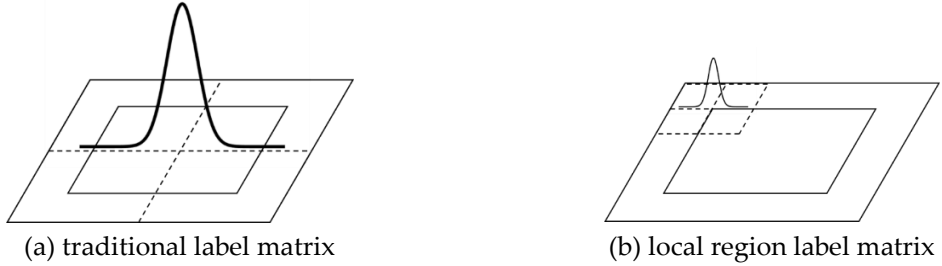

(a) traditional label matrix  (b) local region label matrix

**Figure 4.** The comparison of label matrix between traditional circular samples and local region circular samples in time domain. The curves above the two label matrices denote the two-dimensional Gaussian distribution of both cases.

The comparison between the two-dimensional labels of traditional circular samples and local region circular samples in time domain is shown in Figure 4. The labels in Figure 4(a) are generated for whole searching area, while the labels in Figure 4(b) only considers the correlations of local regions. According to the property of DFT, when two-dimensional label and the self-correlation matrix are transformed from time domain to frequency domain, the high frequency component will move to the origin. Only the matrix elements in the range $[-P, P]$ are considered in the frequency domain. Therefore, the self-correlation kernel matrix and the parameter matrix of the model are greatly reduced.

*3.3 Adaptive model based on multi-feature fusion*

When tracking the target, a single HOG feature focuses on the gradient distribution between every single cell in the target region and less on the local color distribution and texture distribution of the image. Therefore, the local color distribution histogram feature (CH) and the local binary pattern histogram feature (LH) are used as auxiliary features to participate in the training and tracking.

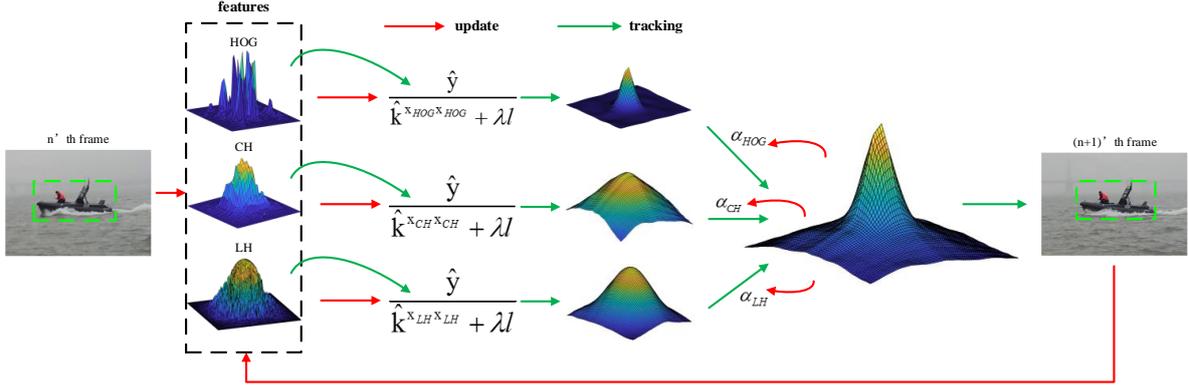

**Figure 5.** Flow chart of proposed model based on adaptive multi-feature fusion. The red arrows indicate that the model parameters are updated according to the predicted results of the current frame. The green arrows indicate that the filtering in this frame is calculated according to the model updated last time and the input image of this frame.

The proposed model based on adaptive multi-feature fusion is described in Figure 5. The red arrows indicate that the model parameters are updated according to the predicted results of the current frame. The green arrows indicate that the filtering in this frame is calculated according to the model updated last time and the input image of this frame. For both color histogram and local binary pattern histogram, a correlation filter model is established respectively. $x_{HOG}, x_{CH}, x_{LH}$ respectively denote the hog feature, color histogram, and local binary pattern histogram. For the same target, the three features are characterized in different fashions, and the filtering responses obtained by the three models are also different. When training the target features in the previous frame, the self-correlation kernel matrix of each feature must be computed first. Then the parameter matrix of each model can be obtained separately. The main model integrates the filtering results of each sub-model and preserves the diversity of each-sub model.

The last filtering response is obtained by weighted summation of the three filter responses on the parameters $\alpha_{HOG}, \alpha_{CH}, \alpha_{LH}$. In the fusing process, the three weights are constantly updated instead of immutable. Actually, the tracking process is a semi-supervised learning process. Only in the first frame, the model knows what the real label is. It not completely accurate to generate the label matrix via the tracking results of each frame in the next update process. Consequently, the three sub-models need to self-evaluate whether the prediction is reliable.

Kullback–Leibler divergence (KL divergence) is a method of measuring the difference between two probability distributions. If two distributions are closer, the KL divergence between will be lower. So we use KL divergence to evaluate the reliability of prediction response.

$$KL(R \| R_{pred}) = \sum_{i,j \in R^{h \times w}} R(i,j) \log \frac{R(i,j)}{R_{pred}(i,j)} \quad (10)$$

In the formula above, $R$ and $R_{pred}$ respectively represents the ideal response and the predicted response. $i, j \in R^{h \times w}$ indicates the coordinates on the two-dimensional matrix. The ideal response $R$ is a two-dimensional Gaussian matrix of which the peak value is at the same location with the peak value of $R_{pred}$. For the three sub-models, 3 KL divergence values can be computed by the formula (6). Then the weights of each sub-model at t'th frame can be calculated as follows:

$$\begin{cases} \eta_{HOG}^t = \dfrac{1}{S^t \times KL_{HOG}^t} \\ \eta_{CH}^t = \dfrac{1}{S^t \times KL_{CH}^t} \\ \eta_{LH}^t = \dfrac{1}{S^t \times KL_{LH}^t} \\ S^t = \dfrac{1}{KL_{HOG}^t} + \dfrac{1}{KL_{CH}^t} + \dfrac{1}{KL_{LH}^t} \end{cases} \quad (11)$$

where $KL_{HOG}^t, KL_{CH}^t, KL_{LH}^t$ are the KL divergence of HOG response, CH response and LH response at t'th frame. $\eta_{HOG}^t, \eta_{CH}^t, \eta_{LH}^t$ are the weights of each sub-model calculated at t'th frame. $S^t$ is a normalized parameter which constrains $\eta_{HOG}^t + \eta_{CH}^t + \eta_{LH}^t = 1$. But they are not the final weights of each sub-model. In order to enhance the robustness of the model and prevent the abnormal value in a frame from causing the model to drift, we adopt the update method to make the weights between the sub-models change smoothly. The following is the updating rule of the final weights:

$$\begin{cases} \alpha_{HOG}^1 = \eta_{HOG}^1 \\ \alpha_{CH}^1 = \eta_{CH}^1 \\ \alpha_{FH}^1 = \eta_{FH}^1 \\ \alpha_{HOG}^t = (1-\lambda) \times \alpha_{HOG}^{t-1} + \lambda \times \eta_{HOG}^t \\ \alpha_{CH}^t = (1-\lambda) \times \alpha_{CH}^{t-1} + \lambda \times \eta_{CH}^t \\ \alpha_{FH}^t = (1-\lambda) \times \alpha_{FH}^{t-1} + \lambda \times \eta_{FH}^t \end{cases} \quad (12)$$

where $\lambda$ is the learning rate of weights $\alpha_{HOG}, \alpha_{CH}, \alpha_{LH}$.

In the first frame, $\eta_{HOG}^1, \eta_{CH}^1, \eta_{LH}^1$ are initialized to satisfy $\eta_{HOG}^t + \eta_{CH}^t + \eta_{LH}^t = 1$. From the formula (8), $\alpha_{HOG}^1 + \alpha_{CH}^1 + \alpha_{LH}^1 = 1$. Afterwards, it can be derived on the basis of recursive relations that $\alpha_{HOG}^t + \alpha_{CH}^t + \alpha_{LH}^t = 1$. Therefore, in the subsequent weights control, the sum of the weights remains constant. The total energy of the final response after fusion is finite, and it always converges with the update of weights.

## 4. Box Regression with Scale Pre-estimation

The self-selection model introduced last section integrates more features to participate in the decision of ship location in the final response. However, the method does not solve the scaling problem of ships. When driving in the camera, the shape and scale of the ships are variable. The size of the bounding box should change in shape when the ship drives near or away from the camera, even when the ship is turning. In this section, a scale pre-estimation method and a box regression method are provided. Firstly, a feature point matching method is used to roughly estimate the scale of the target ship, and then a regression algorithm based on the complete feature is used to precisely adjust the shape of the object.

*4.1 Scale pre-estimation*

Speeded Up Robust Features (SURF) improves the performance on extraction and description of features compared with Scale-invariant Feature Transform (SIFT) algorithm. SIFT descriptors use Gaussian filters to find local extreme according to the Difference of Gaussian (DOG) in images of different scales. By calculating the direction histogram of the local neighborhood of the key point, the direction of the maximal value in the histogram is found as the main direction of the key point. The statistical result of the gradient direction histogram of the key point in the neighborhood Gauss image is the final descriptors. Different from the SIFT descriptor, square filters are adopted in SURF descriptors. The determinant of the Hessian Matrix is used to detect the extreme and integral graph is used to accelerate the operation. It calculates the Haar wavelet transform in the X and Y directions of the pixels around the feature points and takes the maximal value in the sum vector of the two wavelets transform as the direction of the feature points. SURF completes the extraction and description of features in a more efficient way, which can be implemented in real-time in computer vision systems. Like the Sift algorithm, Surf features also have the property of rotation invariance.

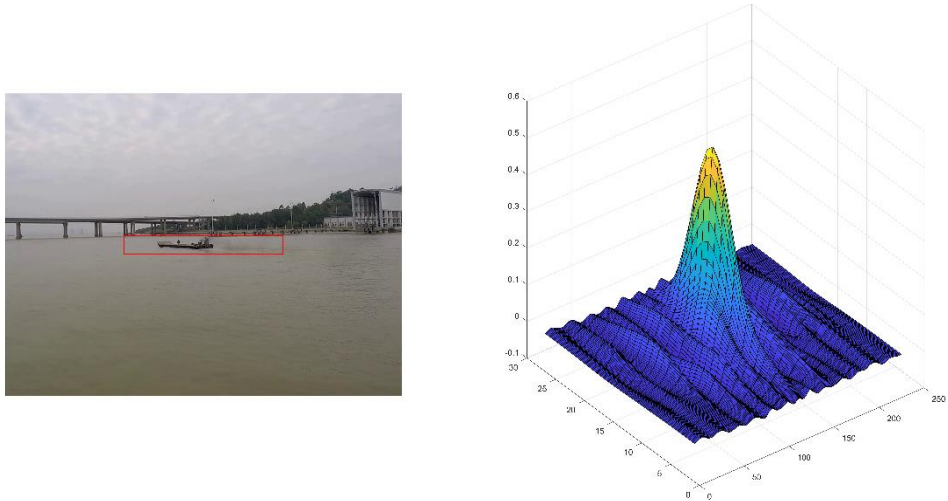

(a) Target away from the camera in the 397th frame

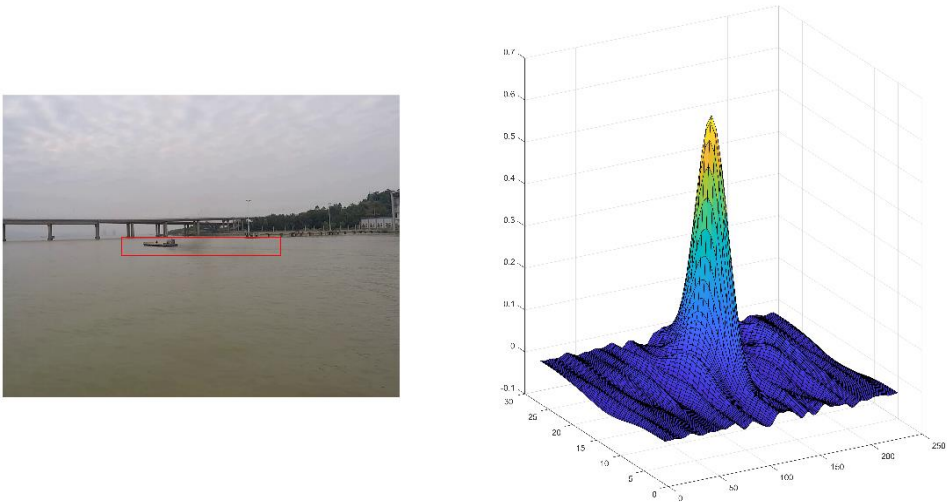

(b) Target away from the camera in the 609th frame

**Figure 6.** Target away from camera causing loss of target: (a) Target away from the camera in the 397th frame; (b) Target away from the camera in the 609th frame. The image on the left is the tracking result of the current frame while the graph on the right is the filtering response of the current frame.

Figure 6 shows a video sequence in which the target is far away from the camera, where (a) and (b) are the image blocks extracted around the target in the 397th frame and 609th frame, respectively. Because the target bounding box cannot automatically adjust the size, so in (a) when the target becomes small in the frame, the bounding box cannot compactly encapsulate the target object. The training samples used in the updating model contain too much background information, causing the drift of the model. This leads to the situation in Figure (b) when the video is at the 609th frame. The filter cannot correctly predict the correct location of the target object while the target is further shrinking. The model will continue to lose the target in next video frames.

Surf algorithm constructs multi-scale pyramid features for image and gets the extreme points by Hessian matrix discriminant. Then, Haar wavelet features in the neighborhood of feature points are counted to determine the main direction. Then the coordinate axis is rotated to the main direction, and the wavelet response in the neighborhood is calculated to generate the feature description vector. Finally, the Euclidean distance between the two feature points is calculated to determine the matching degree.

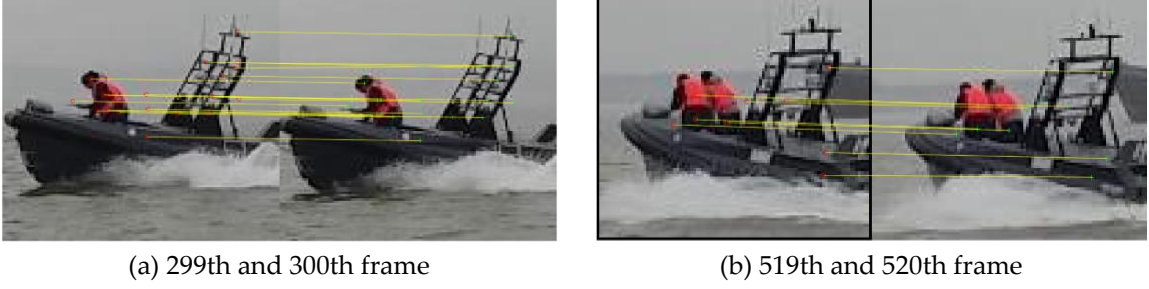

(a) 299th and 300th frame　　　　　　　　　(b) 519th and 520th frame

Figure 7. Schematic diagram of matching points based on surf feature in adjacent frames: (a) 299th and 300th frame; (b) 519th and 520th frame.

This paper adopts a key point matching strategy based on surf features for scale pre-estimation. Figure 7 shows the matching points using surf features between two adjacent frames. Because there is no big jitter on camera and target object will not have large deformation, it is easier to match as many key points as possible in adjacent frames. The scaling ratio of the target can be calculated by the minor changes between these key points. Because the bounding box contains a small amount of background information, the feature points on the non-target objects will be matched. Therefore, some extra conditions are needed to enhance the contribution of the feature points belong to the object part and weaken the contribution of the feature points in the background area.

Assume that the key points detected in the previous frame are $K^p = \{K_1^P, K_2^P, \ldots, K_M^P\}$, the key points detected in the latter frame are $K^l = \{K_1^l, K_2^l, \ldots, K_N^l\}$. Set of key matches for success matching：

$$Pairs = \{(K_1^p, K_1^l), (K_2^p, K_2^l), \ldots, (K_T^p, K_T^l)\}, T \leq M, T \leq N \tag{13}$$

where $T$ is the number of matching pairs, $K$ is a coordinate of a key point.

Each key point is assigned different weights in accordance with their importance. It is assumed that the key points near the target center are more likely to be part of the target. Therefore, the points close to the target center should be given higher weights, and those far away from the target center will be given lower weights. Therefore, different weights are assigned to the key points according to the filter response values at the position of each key point. Assume that the weight at $K_i^p$ is $\omega_i^p$ and the weight at $K_j^l$ is $\omega_j^l$. The weighted centroid of the key points can be calculated through the weights in the previous and latter frames:

$$M^p = \frac{\sum_i \omega_i^p K_i^p}{\sum_i \omega_i}, M^l = \frac{\sum_j \omega_j^l K_j^l}{\sum_j \omega_j} \tag{14}$$

where $M^p$ and $M^l$ are the centroids of key points in the previous frame and the latter frame.

The pre-estimated scale can be calculated from the distance between the centroid of the feature point and the target center：

$$scale(p,l) = \frac{\|M^l - C^l\|}{\|M^p - C^p\|} \tag{15}$$

where $C^p$ and $C^l$ are the center points of the target in the previous frame and the latter frame.

*4.2 A bounding box regression method*

Scale variation range of the target object can be estimated by key points matching method mentioned in the previous paragraph. The purpose of the estimation is to cope with changes in the distance between the target and the camera. In addition to the scale changes, there is occlusion or morphological changes between adjacent frames. A complementary approach should be used to fine-tune the predicted bounding box to make it more precise.

For our ship tracking task, a bounding box regressor is specially trained. Suppose that $N$ pairs of samples $\{(S^i, R^i)\}_{i=1}^{N}$ participate in the training process. Each pair of samples represents a real bounding box $R^i = (R_x^i, R_y^i, R_w^i, R_h^i)$ of the target and a sampled box $S^i = (S_x^i, S_y^i, S_w^i, S_h^i)$ tightly around $R^i$. $(S_x^i, S_y^i)$ and $(S_w^i, S_h^i)$ respectively indicate the center coordinate and the width/height of the sampled box. Similarly, $(R_x^i, R_y^i)$ and $(R_w^i, R_h^i)$ respectively represent the center coordinate and the width/height of the ground-truth bounding box of the target. The purpose of training the regressor is to get an ideal mapping function from the predicted bounding box to the ground-truth bounding box. The relationship between the predicted box and the real box is shown in Figure 8:

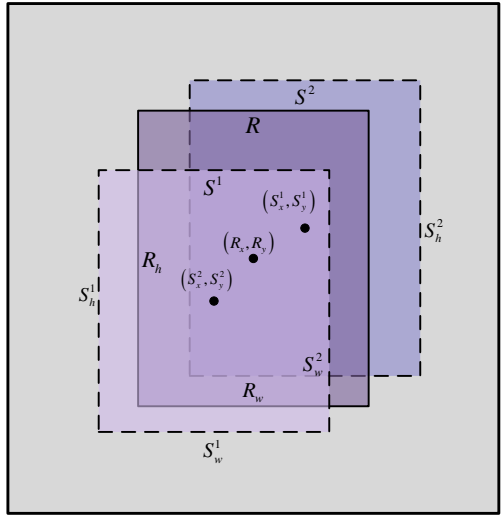

**Figure 8.** The relationship between the sampled box $S$ and the real box $R$. After training the regressor by the mapping function from a real bounding box to the sampled bounding boxes. A predicted bounding box can be generated by the regressor in the testing stage.

For the four components $x, y, w, h$ of a bounding box, a regressor for predicting the feature $F$ is defined:

$$\begin{cases} \sigma_x = \mathrm{w}_x^{\mathrm{T}} F_{HOG}(S) \\ \sigma_y = \mathrm{w}_y^{\mathrm{T}} F_{HOG}(S) \\ \sigma_w = \mathrm{w}_w^{\mathrm{T}} F_{HOG}(S) \\ \sigma_h = \mathrm{w}_h^{\mathrm{T}} F_{HOG}(S) \end{cases} \quad (16)$$

where $F_{HOG}(S)$ indicates the HOG features extracted from the box $S$. Four parameter vectors of linear regressor based on HOG feature are defined as $\mathrm{w}_x, \mathrm{w}_y, \mathrm{w}_w, \mathrm{w}_h$. Four outputs $\sigma_x, \sigma_y, \sigma_w, \sigma_h$ are used to approximate four corresponding target parameters defined as:

$$\begin{cases} t_x = (R_x - S_x)/S_w \\ t_y = (R_y - S_y)/S_h \\ t_w = \log(R_w/S_w) \\ t_h = \log(R_h/S_h) \end{cases} \quad (17)$$

where $(\sigma_x, \sigma_y)$ is the relative offset from sampled box to the real box output by the regressor. $(t_x, t_y)$ are the ideal relative offset between sampled box to the real box during training. $(\sigma_w, \sigma_h)$ are the relative scale ratio under a log function output by the regressor. $(t_w, t_h)$ are the ideal relative scale under a log function.

The problem can be divided into 4 ridge regression problems:

$$\begin{cases} w_x = \arg\min_{\hat{w}_x} \sum_i^N \left(t_x^i - \hat{w}_x^T F_{HOG}(S^i)\right)^2 + \lambda \|\hat{w}_x\|^2 \\ w_y = \arg\min_{\hat{w}_y} \sum_i^N \left(t_y^i - \hat{w}_y^T F_{HOG}(S^i)\right)^2 + \lambda \|\hat{w}_y\|^2 \\ w_w = \arg\min_{\hat{w}_w} \sum_i^N \left(t_w^i - \hat{w}_w^T F_{HOG}(S^i)\right)^2 + \lambda \|\hat{w}_w\|^2 \\ w_h = \arg\min_{\hat{w}_h} \sum_i^N \left(t_h^i - \hat{w}_h^T F_{HOG}(S^i)\right)^2 + \lambda \|\hat{w}_h\|^2 \end{cases} \quad (18)$$

where $\lambda$ is a regularization parameter.

The training of the regressor is off-line, and the batch gradient descent (BGD) method is used to solve the optimization problem above. The off-line means the regressor is firstly trained on several videos, and then be fine-tuned at the initial frame of the tracking video. At each iteration, 8 bounding boxes around the actual target bounding box are sampled and the parameters $w_x, w_y, w_w, w_h$ are updated once. After several iterations, the regressor converges to a locally optimal result on the training set.

When the model tracks the target, the input of the regression is the bounding box $P^{pre} = \left(P_x^{pre}, P_y^{pre}, P_w^{pre}, P_h^{pre}\right)$ predicted by the scale prediction method. According to the approximation relation between $\sigma_x, \sigma_y, \sigma_w, \sigma_h$ and $t_x, t_y, t_w, t_h$ in formula (12) (13), the final position $P^{final} = \left(P_x^{final}, P_y^{final}, P_w^{final}, P_h^{final}\right)$ can be obtained.

$$\begin{cases} P_x^{final} = P_x^{pre} + w_x^T F_{HOG}\left(P^{pre}\right) \times P_x^{pre} \\ P_y^{final} = P_y^{pre} + w_y^T F_{HOG}\left(P^{pre}\right) \times P_y^{pre} \\ P_w^{final} = P_w^{pre} \exp\left(w_w^T F_{HOG}\left(P^{pre}\right)\right) \\ P_h^{final} = P_h^{pre} \exp\left(w_h^T F_{HOG}\left(P^{pre}\right)\right) \end{cases} \quad (19)$$

In addition, when selecting training pairs, the overlap ratio between the sampled box and the real box greatly affects the performance of the regression. The overlap ratio represents the ratio of the overlapping area to the merging area of two bounding boxes. If the predicted overlap ratio between the predicted box and the real box is large enough, then the fine-tuning is valid. Conversely, it is difficult for the regressor to predict the accurate location.

## 5. Experiments

The experimental environment in this paper is CPU: Intel Core i7-4790 3.6GHz; memory: 16GB, without using GPU to take part in the operation. The evaluation software is Matlab R2017a. The video data of the evaluation is provided by the large data project of intelligent transportation in our laboratory. 10 video sequences have been intercepted with different lengths. These videos have a frame rate of 25 FPS and a resolution of $1920 \times 1080$. The dataset includes the videos captured by the sea and the videos recorded on a traveling ship. Most of the targets in these videos are the ships sailing into or out of the bay. By tracking the course of these ships, we can judge the relative position of the ship and the restricted area, and analyze the behavior of the ship. The performance comparison between the BRCF and KCF as well as DSST on the dataset in this paper is shown below.

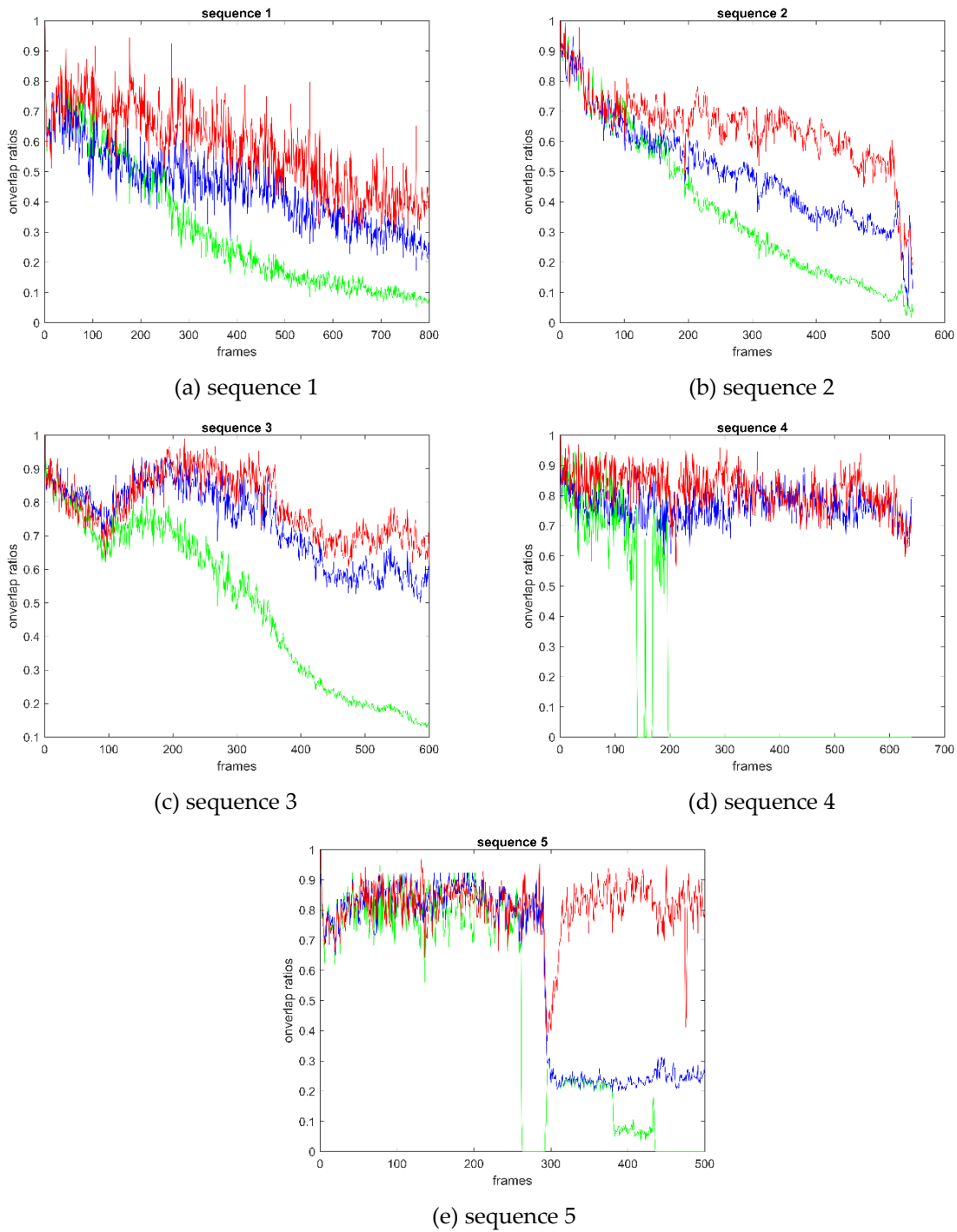

**Figure 9.** Curve of overlap ratio between predicted box and real box: (a) sequence 1; (b) sequence 2; (c) sequence 3; (d) sequence 4; (e) sequence 5. The points on the curves indicate the overlap ratio between the predicted bounding box and the ground-truth bounding box.

Figure 9 shows the overlap ratio curve of video sequence 1-5. The points on the curves indicate the overlap ratio between the predicted bounding box and the ground-truth bounding box. From the overall curve, BRCF is significantly higher than KCF in each video sequence. In the curve of sequence 1 and 2, the holistic overlap ratio of BRCF is higher than that of KCF and DSST. At the beginning of Sequence 2, the overlap ratio of BRCF and DSST was close, but the curve of DSST in the latter part was obviously lower than that of BRCF. In sequence 3, BRCF is almost equivalent to DSST at the initial 200 frames. Then BRCF is a little superior to DSST in the next hundreds of frames. Similarly, in sequence 4, BRCF performs slightly better than DSST in the first 300 frames and then has the same effect as DSST. In sequence 5, BRCF has comparable performance with DSST and KCF at the beginning 300 frames. After an interference at about the 300'th frame, DSST and KCF lose the target synchronously while the proposed BRCF method still keeps tracking of the target ship in next frames.

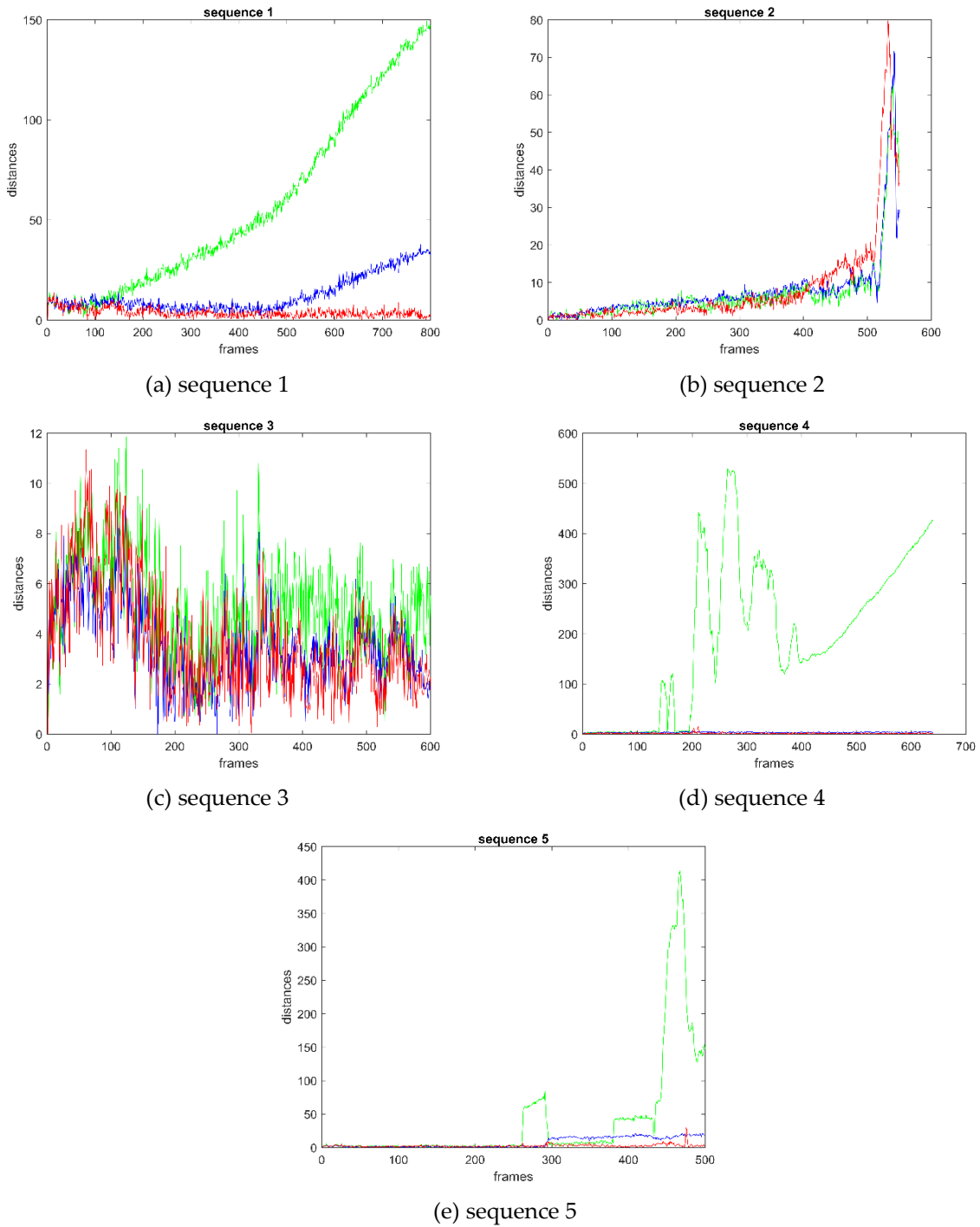

(a) sequence 1

(b) sequence 2

(c) sequence 3

(d) sequence 4

(e) sequence 5

**Figure 10.** Curve of distance between predicted box and real box: (a) sequence 1; (b) sequence 2; (c) sequence 3; (d) sequence 4; (e) sequence 5; (f) sequence 6; (g) sequence 7. The points on the curves denote the distance between the center of the predicted bounding box and the ground-truth bounding box.

Figure 10 shows the distance curve of video sequence 1-5. The points on the curves denote the distance between the center of the predicted bounding box and the ground-truth bounding box. Except for sequence 2, the distance of BRCF is obviously lower than that of KCF on any other sequence. The performance of BRCF and DSST is close to each other in sequence 3 and sequence 4. The performance of BRCF is slightly better than DSST after the 300'th frame of sequence 5. In sequence 1, the BRCF performs better than the other two methods on distance curve. Comparing the overlap ratio curve with the distance curve, it can be inferred that the overall overlap ratio of BRCF is higher than that of KCF and DSST, and the overall distance between the predicted bounding box and the ground-truth bounding box is lower than that of KCF and DSST. It shows that the bounding box predicted by BRCF is closer to the ground-truth bounding than that predicted by KCF and DSST.

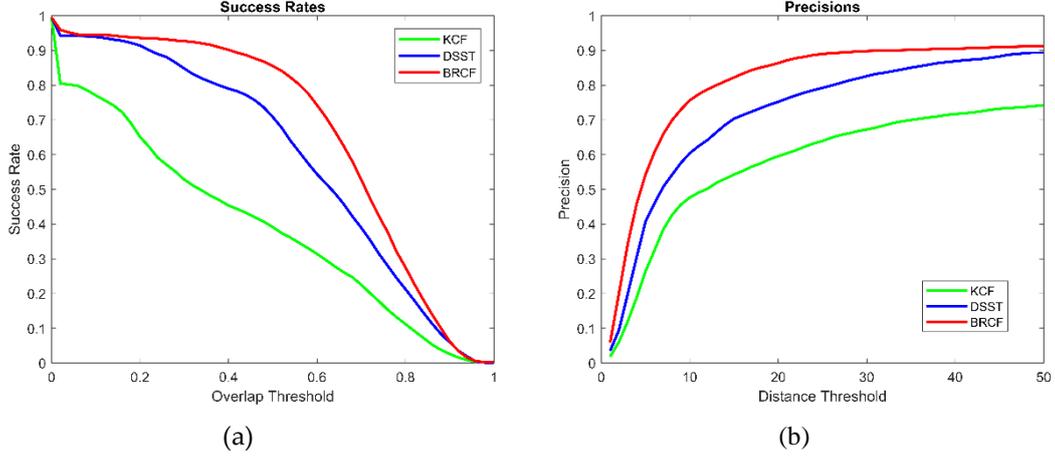

**Figure 11.** The performances of KCF, DSST, and BRCF on all video sequences: (a) The curve of success rate over the overlap ratio threshold; (b) The curve of precision over distance threshold. The success rate is the ratio of the number of successful frames to the total number of frames in all video sequences. Distance is calculated by the Euclidean distance between the center pixels of two bounding boxes.

Figure 11 (a) shows the curve of success rate over the overlap ratio threshold of BRCF and KCF as well as DSST. The tracking success is defined when the overlap ratio of the predicted bounding box and the ground-truth bounding box is higher than a certain threshold value. The success rate is the ratio of the number of successful frames to the total number of frames in all video sequences. Within all overlap ratio thresholds, the success rate of BRCF is much higher than that of KCF. The performance of BRCF is better than that of DSST in the most threshold range. This shows that the BRCF method is more effective than the DSST method in dealing with the scale change.

Figure 11 (b) shows the curve of precision over the distance threshold of BRCF and KCF as well as DSST. Distance is calculated by the Euclidean distance between the center pixels of two bounding boxes. When the distance of the predicted bounding box and the ground-truth bounding box is lower than a certain threshold value, the tracking of this frame is precise. The precision is the ratio of the number of precise frames to the total number of frames in all video sequences. It can be seen from the figure that the precision of the BRCF is higher than that of the DSST and KCF in all scale ranges.

Table 1 shows the comparison of total performance on average overlap ratio, average distance, average success rate and average precision. The average overlap ratio and average success rate of BRCF are higher than that of DSST by over 0.08 respectively. The average distance of BRCF is lower than that of DSST by over 8-pixel length. The average precision of BRCF is higher than that of DSST by over 0.08. In summary, the total performance of BRCF is better than DSST on our dataset.

Table 2 shows the comparison of the processing rate on our marine traffic dataset. The average processing rate of KCF, DSST and BRCF are 131.80 FPS, 23.07 FPS and 44.98 FPS. KCF reduces video scale and only extracts the gray HOG features of the image, so the processing speed is faster. The difference between BRCF and DSST is that no redundant correlation filter is used to deal with the scale, only the regressor is used to adjust the scale without extracting redundant HOG features. In addition, the local region mining method reduces some parameters of the model.

Table 3 shows a comparison of time consumption on scale calculation between DSST and BRCF. DSST needs 0.009s for scale prediction, 0.012s for scale training. BRCF only needs 0.008s for scale prediction.

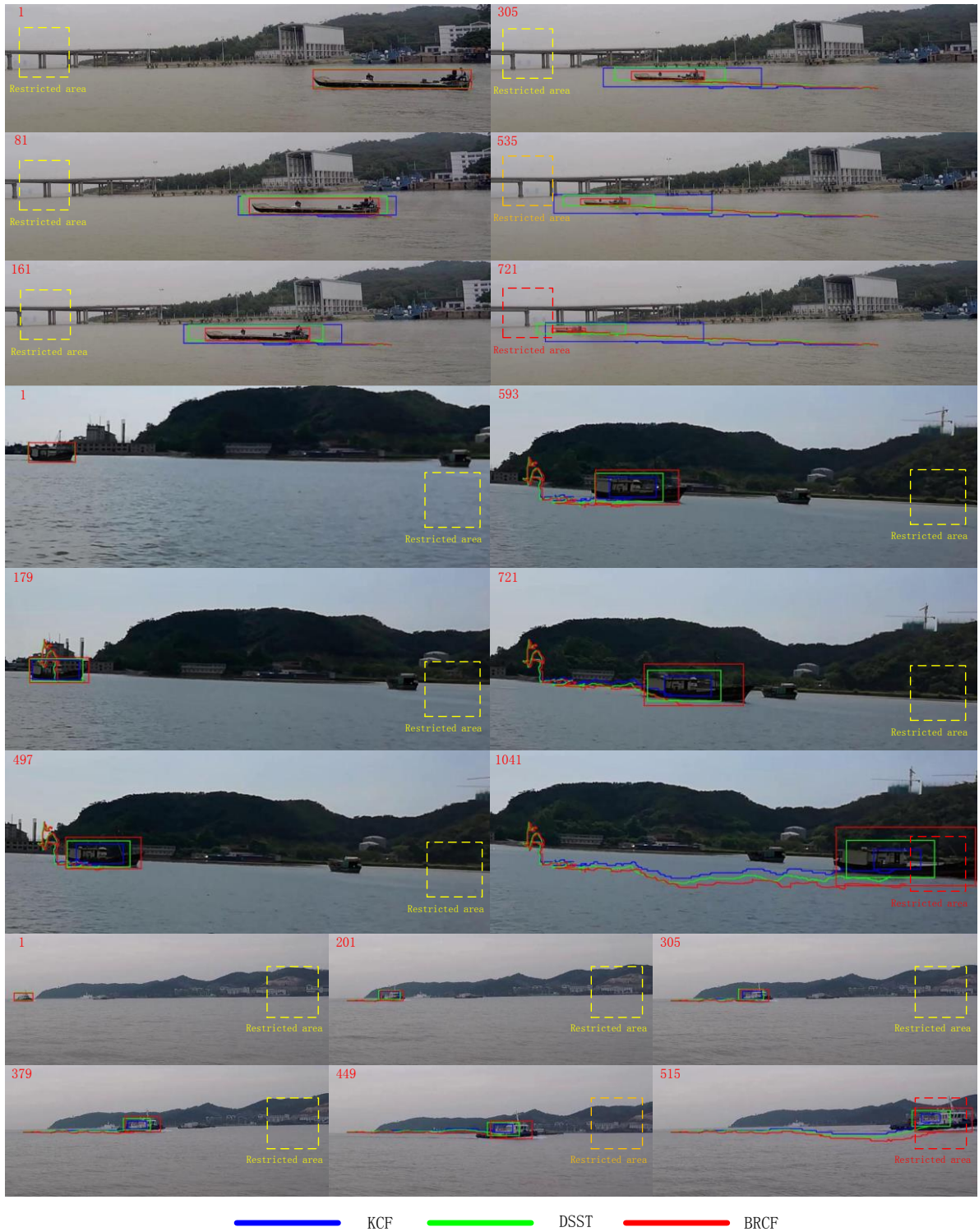

**Figure 12.** The detection of abnormal events on the sea by the three tracking methods. When the ships sailing near the preset restricted area, the smart ocean system will give an alarm. The color of the dashed rectangle reflects the distance between the moving ship and the restricted area.

**Table 1.** Comparison of three methods on the average overlap ratio, the average distance, the average success rate, and the average precision

|  | KCF | DSST | **BRCF** |
|---|---|---|---|
| Average overlap ratios | 0.3881 | 0.5871 | 0.6652 |
| Average distances | 80.3250 | 28.4143 | 19.8536 |

| Average success rates | 0.3922 | 0.5860 | 0.6624 |
|---|---|---|---|
| Average precisions | 0.5780 | 0.7204 | 0.8059 |

**Table 2.** Comparison of processing rate on our marine traffic dataset

| Methods | The performance of the three algorithms on 20 videos | | | | | Average rate |
|---|---|---|---|---|---|---|
| KCF | 163.12 | 136.47 | 133.62 | 128.32 | 89.71 | 131.80 |
| | 166.39 | 137.35 | 241.94 | 176.83 | 91.28 | |
| DSST | 21.44 | 15.63 | 13.96 | 18.26 | 8.61 | 23.07 |
| | 18.13 | 10.25 | 9.68 | 7.84 | 6.87 | |
| BRCF | 40.97 | 39.85 | 32.74 | 39.51 | 23.88 | 44.98 |
| | 46.89 | 52.72 | 29.58 | 26.86 | 31.79 | |

**Table 3.** Comparison of time consumption on scale calculation

| | DSST | | **BRCF** | |
|---|---|---|---|---|
| Time consumption(s) | Scale training | Scale prediction | Scale training | Scale prediction |
| | 0.012 | 0.009 | **0** | **0.008** |

The detection of abnormal events on the sea by the three tracking methods of KCF, DSST and BRCF on the three video sequences are shown in Figure 12. In the Smart Ocean System, one of the most important tasks is to detect and track abnormal events and illegal ships on the sea. A restricted area is set in advance for anomaly detection. When the ships sailing near the preset restricted area, the system will give an alarm. The color of the dashed rectangle reflects the distance between the moving ship and the restricted area. As the ship moves closer to the restricted area, the color of the rectangular box becomes darker until the alarm is triggered. The curve connected to the bounding box represents the tracking trajectory of each algorithm for the target ship. The red number in the upper left corner of each frame indicates the serial number in the video sequences. It can be seen that the scale of the target box of KCF method cannot adjust the box size when the object size changes. The bounding box of DSST can adjust its size according to the change of object scale, but it cannot change the aspect ratio of the box. Once the object is deformed, the rectangle box will not be able to wrap the object tightly. The proposed BRCF method is more adaptable to the deformation of objects and other scenes. It can adjust the scale freely following the shape change of objects.

## 6. Conclusions

In this paper, we proposed a self-selective CF model based on box regression. In view of the severe boundary effect, the range of positive and negative samples is controlled according to the circular shift distance. The diversity of features is improved by a self-selective model with multi-response fusion. Combining the key points matching method for scale pre-estimation of the tracking target, the regression method can make the bounding box change in accordance with arbitrarily shaped targets. The experimental results show that the average success rated and precisions were higher than DSST by about 8 percentage points in the laboratory of marine traffic dataset. In terms of processing speed, the proposed method is higher than DSST by nearly 22 FPS. It can achieve almost real-time processing. Meanwhile, the proposed method can effectively deal with the problem of object size changes and background interference.